\documentclass[letterpaper, 10 pt]{ieeeconf}
\IEEEoverridecommandlockouts 
\usepackage{mathrsfs}
\usepackage{dutchcal}
\usepackage{amsmath,amssymb,amsfonts}
\usepackage{graphicx}
\usepackage{textcomp}
\usepackage{xcolor}
\usepackage{soul}
\usepackage{hyperref}
\usepackage{url}
\usepackage{gensymb}
\usepackage{adjustbox}
\usepackage{booktabs}
\usepackage{wrapfig}
\usepackage{multirow}
\usepackage{rotating}
\usepackage{changes}

\usepackage{subfigure} 
\usepackage{caption}
\usepackage{cleveref}
\crefname{figure}{Fig.}{Figures}
\crefname{section}{Sec.}{Sections}
\crefname{equation}{Eq.}{Equations}
\crefname{table}{Table.}{Tables}
\usepackage{algorithm}
\usepackage{algpseudocode}

\usepackage[sorting=none,style=numeric,backend=biber]{biblatex}
\title{
{
Imitation Learning from Human Motion Alone Does Not\\ Guarantee Biomechanically Plausible Gait Kinetics
}
}

\author{
Xinyi Liu$^{1}$, Jangwhan Ahn$^{1}$, Edgar Lobaton$^{2}$, Jennie Si$^{3}$, He Huang$^{1}$ 
\thanks{$^{1}$ Joint Department of Biomedical Engineering, University of North Carolina - Chapel Hill, Chapel Hill, NC,  USA
{\tt\small Email: xinyili@ad.unc.edu, jahn26@ncsu.edu, hehuang@email.unc.edu}}
\thanks{$^{2}$ Department of Electrical and Computer Engineering, North Carolina State University, Raleigh, NC,  USA
{\tt\small Email: ejlobato@ncsu.edu}}
\thanks{$^{2}$ School of Electrical, Computer and Energy Engineering, Arizona State University, Tempe, AZ, USA
{\tt\small Email: si@asu.edu}}
}

\begin{document}
\maketitle
\thispagestyle{empty}
\pagestyle{empty}

\begin{abstract}
{

Imitation learning (IL) via motion data is increasingly used in robotics and character animation, yet its ability to be applied to human biomechanics to estimate physiologically plausible joint moments without explicit kinetic information remains unclear. 
In this study, we examined whether motion imitation alone can estimate reasonable biological joint moments in human walking. 
We compared motion-only IL (MOIL) against a kinetics-aware IL (KAIL) framework, which incorporates additional ground reaction forces (GRFs) and center of pressure (CoP) in imitation rewards. A progressive kinetic reward shaping approach was used to examine the contribution of each kinetic term.
Experiments were conducted using walking data from two non-disabled participants at three speeds (0.9, 1.2, and 1.5 m/s). While both MOIL and KAIL achieved comparable kinematic tracking accuracy, 
KAIL produced kinetics more closely aligned with biomechanical values across participants and walking speeds, reducing the root mean square errors (RMSEs) of vertical GRF and CoP by 40\% and 48\%, and of hip, knee, and ankle moments by 19\%, 18\%, and 35\%, respectively, relative to MOIL. 
These findings highlight a fundamental limitation of MOIL approaches, as their failure to accurately estimate human-like gait kinetics may lead to erroneous interpretations of gait biomechanics and adversely affect downstream applications such as wearable robot design and the diagnosis of pathological gait.}

\end{abstract}

\textit{\textbf{Index Term---}} gait biomechanics, reinforcement learning, imitation learning 


\section{INTRODUCTION}
\label{sec:introduction}
Imitation learning (IL), also known as learning from demonstrations, is a machine learning paradigm in which an agent learns a policy by observing demonstrations from an expert. The goal of IL is to reproduce similar input–output behaviors \cite{argall2009survey}.  In the reinforcement learning (RL) formalism, IL typically seeks a policy that maximizes expected return with a reward function that explicitly encourages similarity to expert trajectories \cite{zhang_reinforcement_2021}. It offers a promising approach to 
replicate demonstrated behaviors in forward dynamics simulations \cite{Zare2023ilReview}. 
Recently, the rise of deep RL has enabled scalable policy representations that can imitate complex behaviors from raw sensory inputs. RL-based IL has significantly advanced applications across various domains, including robotic manipulation \cite{zhao2025resmimicgeneralmotiontracking, zhu2018reinforcementimitationlearningdiverse, ke2021grasp}, physics-based animation \cite{deepmimic2018, yuan_residual_2020}, autonomous driving \cite{chen2019autodriving, li2024safelearning}, humanoid control \cite{ yang_omniretarget_2025, 2025humanoidreview} and assistive systems \cite{fang2025, joseph2023}, among others. 
 \begin{figure}
    \centering
    \includegraphics[width=0.95\linewidth]{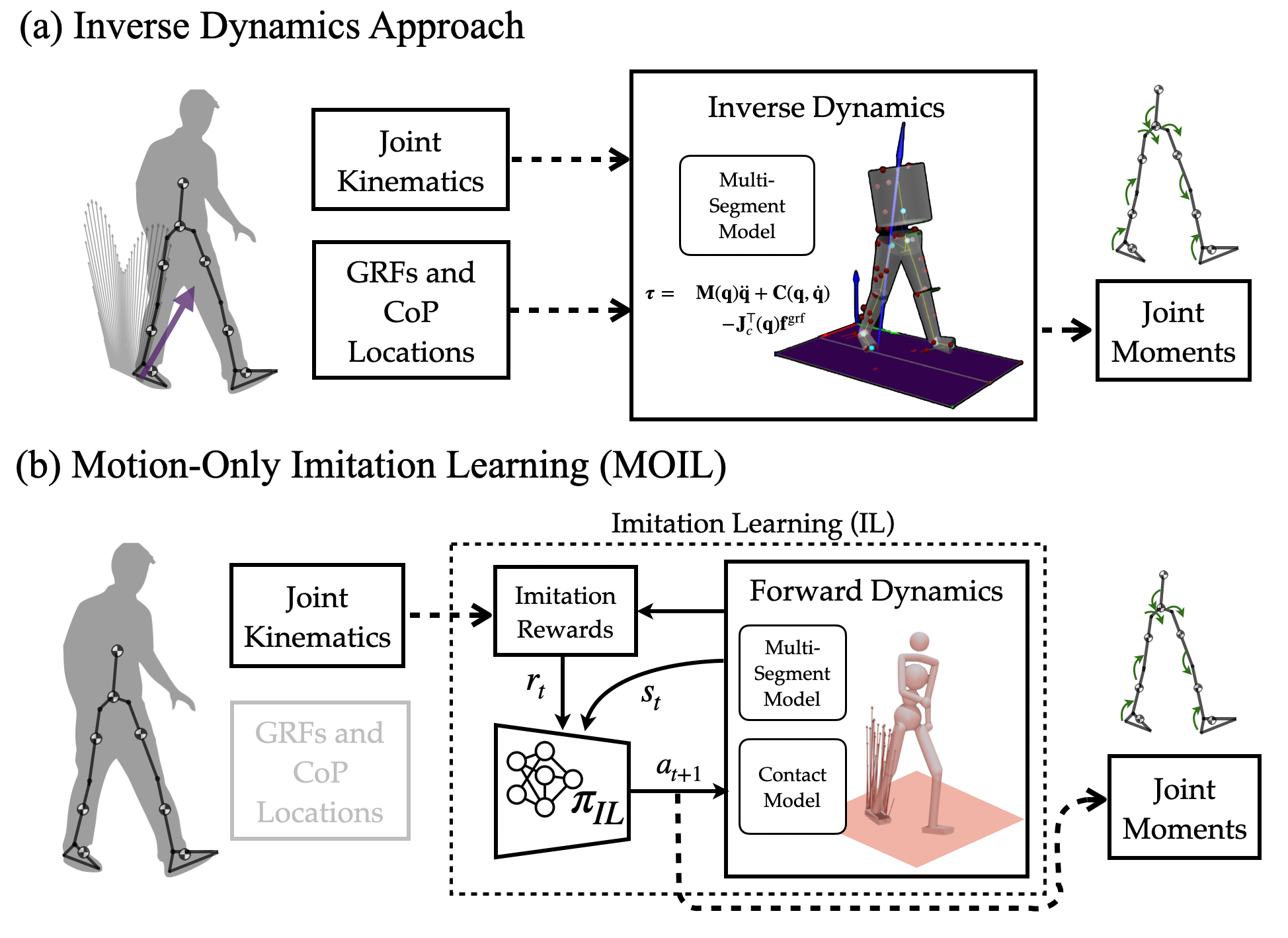}    
    \caption{Comparison of (a) a biomechanics inverse dynamics pipeline and (b) a motion-only imitation learning (MOIL) pipeline integrating a reinforcement learning agent with forward dynamics; ground reaction force (GRF) and center of pressure (CoP) measurements are omitted in MOIL approaches. 
    }
    \label{fig:intro}
\end{figure}

RL-based IL has recently attracted the attention of researchers in human biomechanics and wearable robotics, since IL policies can reproduce expert-demonstrated motions by generating the corresponding joint moments and muscle activation patterns required for smooth, natural motion \cite{deepmimic2018, yuan_residual_2020}.  In this study, we refer to these IL pipelines as \textit{motion-only} \textit{IL} (MOIL). 
Historically,
classic biomechanics analysis provides joint moment estimates from kinematics via inverse dynamics \cite{lynch2017modern}. 
For locomotive tasks, given an approximate multi-segment dynamical model, 
joint trajectories obtained via inverse kinematics \cite{Cotton_Diff_2025} from motion capture data, and measured ground reaction force (GRF) signals from force plates or instrumented treadmills, one can apply the Recursive Newton–Euler algorithm to compute biological joint moments,  
as shown in \cref{fig:intro}(a).  Often, specialized instruments are required in the gait laboratory to collect these data for inverse dynamics calculation. 
In contrast,  MOIL tracks joint kinematics or sparse task rewards without relying on external force measurements such as GRFs, as shown in \cref{fig:intro}(b). This approach is particularly appealing because measuring external forces is difficult in real-world and clinical settings, whereas human motion can now be measured ubiquitously, driven by rapid progress in wearable sensors and computer vision \cite{das2023comparison, uhlrich2023opencap}.  
MOIL may enable biomechanists, clinicians, and researchers to estimate joint {moments} and muscle activities without relying on specialized force-sensing equipment, which is typically confined to gait laboratories \cite{hulleck2022present}. This, in turn, could enable richer cause-and-effect biomechanics analyses \cite{song2021deepRL} and what-if analyses of responses to wearable robotic devices \cite{bianco2023simulating, wen2017}, support clinically interpretable mobility assessments and diagnostic tools beyond the lab \cite{cotton_kintwin_2025, schumacher2023deprl}, and facilitate co-design of shared-control wearable systems that generalize outside controlled clinical environments \cite{van_der_kooij_ai_2025}.


 {Despite its potential, it remains unclear whether MOIL ensures the production of human-like joint moments in walking. This is because human gait dynamics are largely influenced by foot-ground interactions, where the reaction forces and their applied points on the foot affect the system's dynamics and shape the joint kinetics \cite{winter2009biomechanics}.
  Without explicit constraints on these foot-ground contact interactions, MOIL may explore and learn foot-ground contact patterns different from those in real gait, and therefore influence the estimation of biological joint moments.} 
 Although recent works have reported promising forward dynamics simulations of natural human gait using variants of MOIL \cite{schumacher2023deprl, al-hafez_locomujoco_2023}, these studies rarely provide a systematic evaluation of the resulting GRFs and joint moments against experimental GRF measurements and inverse dynamics estimates. Furthermore, in cases where kinetics were reported \cite{ cotton_kintwin_2025, chiu2025speedadaptive}, the correspondence with inverse dynamics {joint} moment and GRF profiles is often incomplete or inconsistent across joints, tasks, or studies. {Finally, another reason to investigate MOIL is the growing interest in its applications within biomechanics. Inaccurate joint moment estimates can further lead to erroneous estimates of muscle forces and activations, which may compromise the  applications of MOIL in clinical diagnosis, rehabilitation strategy development, or exoskeleton control policy design. }

 

{Hence, the objective of this study is to investigate whether MOIL is sufficient to learn policies that produce biomechanically plausible joint moments in walking. We argue that additional kinetics-related measurements, such as foot contact forces, remain essential for estimating human-like gait kinetics. To address the question, we extended MOIL \cite{yuan_residual_2020} to \textit{kinetics-aware IL} (KAIL), which adds rewards for matching GRFs and CoP during normal walking. From walking data  {of two} non-disabled {participants}, we trained {subject-specific }policies under different reward configurations with a progressive
kinetic reward shaping approach. We then assessed the learned policies by comparing simulated GRFs, CoP, and joint moments against those derived from inverse dynamics as references. The outcome of this study may inform future biomechanical applications of IL}\footnote{ Our source code, dataset, and simulation environments will be made publicly available upon acceptance of the manuscript.}.

{\cref{sec:prelim} introduces the dynamic model used for forward dynamics simulation. \cref{sec:methodology} formalizes the IL framework and reward design. \cref{sec:experiments} describes the human and simulation experiments. \cref{sec:results} presents the learning performance and biomechanical evaluation results. Finally, \cref{sec:discussion} discusses and concludes our main findings.}

\section{HUMAN DYNAMICS MODEL}
\label{sec:prelim}

\label{subsec:dynamics} 
\begin{figure}
    \centering
    \includegraphics[width=0.8\linewidth]{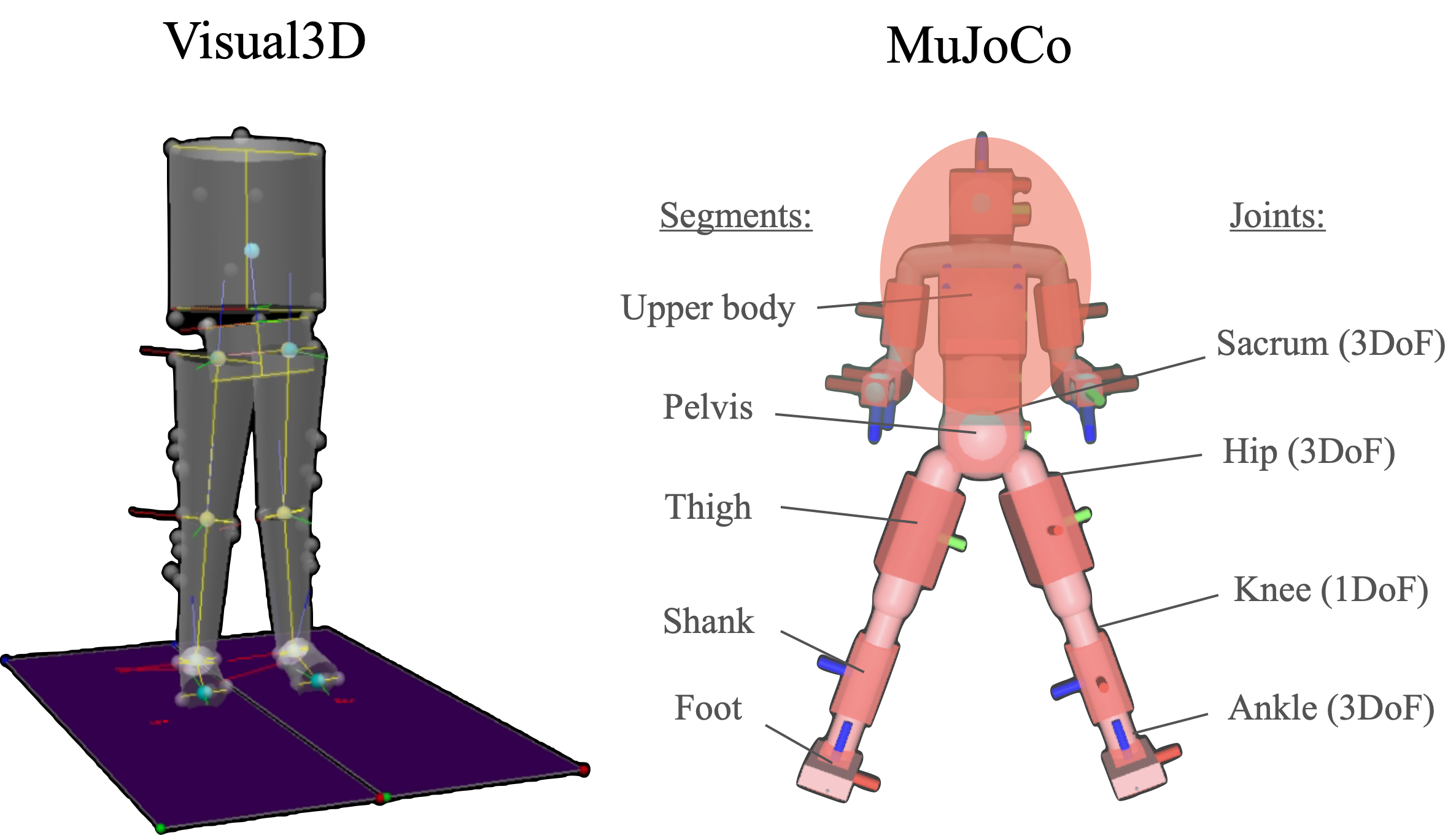}
        \caption{Multi-segment rigid body models used for the inverse dynamics and imitation learning. The Visual3D inverse dynamics model and the MuJoCo forward dynamics model shared identical anthropometric configurations (segment lengths, masses, and inertial properties) to ensure consistency.}
    \label{fig:dynamics}
\end{figure}

The human body is modeled as a floating-base, multi-segment system, as illustrated in \cref{fig:dynamics}. The system configuration is parameterized by the generalized coordinates $\mathbf{q} = [\mathbf{p}, \mathcal{q}, \boldsymbol{\psi}] ^{\top}$, where $\mathbf{p} \in {\mathbb{R}^{3}}$ denotes the global root (pelvis) position, $\mathcal{q} \in \mathbb{R}^4$ represents the root orientation quaternion and $\boldsymbol{\psi} \in \mathbb{R}^{17}$ defines the joint space coordinates. Specifically, the trunk is modeled as a single rigid segment connected to the root via a 3-DoF spherical joint, while each lower limb consists of a 3-DoF hip, a 1-DoF knee (flexion/extension), and a 3-DoF ankle joint. 
Forward dynamics simulations are implemented in MuJoCo \cite{Todorov2012Mujoco}, following the equations of motion for a constrained floating-base system, 
\begin{equation}
\label{eq:fd}
    \mathbf{M}(\mathbf{q})\,\ddot{\mathbf{q}} + \mathbf{C}(\mathbf{q}, \dot{\mathbf{q}}) = \left[ \begin{matrix}
    \boldsymbol{\xi} \\
    \boldsymbol{\tau} 
    \end{matrix} \right]
    + \mathbf{J}_c(\mathbf{q})^\top \mathbf{f}^{\mathrm{grf}},
\end{equation}

where $\dot{\mathbf{q}}$ and $\ddot{\mathbf{q}}$ are the generalized velocity and acceleration, \(\mathbf{M}(\mathbf{q})\) is the generalized mass–inertia matrix, \(\mathbf{C}(\mathbf{q}, \dot{\mathbf{q}})\) collects Coriolis, centrifugal, and gravitational terms, \(\boldsymbol{\tau} \in \mathbb{R}^{17}\) denotes the actuated joint torques and $\boldsymbol{\xi} \in  \mathbb{R}^{6}$ represents the corrective residual wrench applied to the floating base {that compensates for model mismatch~\cite{yuan_residual_2020}, as residuals do in inverse dynamics~\cite{hicks2015my}}.
{Ground contact is handled by MuJoCo's soft-constraint contact model, which computes the contact forces at each time step by solving a convex optimization problem; $\mathbf{J}_c(\mathbf{q})$ is the contact Jacobian and $\mathbf{f}^{\mathrm{grf}}$ stacks the resulting contact forces at the feet.}
Foot contact is modeled as four discrete contact points for each side, and a contact layer is implemented via the default contact margin 
to obtain reasonably smooth foot–ground interactions. 

\section{LEARNING KINETICS}
\label{sec:methodology}

\subsection{Problem Formulation} 
The imitation learning (IL) problem is formulated as a finite-horizon Markov decision process (MDP) over a multi-segment dynamics model (see \cref{sec:prelim}), following the Residual Force Control framework \cite{yuan_residual_2020}, as shown in \cref{fig:ppo_schematic}.

\subsubsection{MDP Definition} 

The RL environment defines an MDP \(\mathcal{M} = (\mathcal{S},\mathcal{A},P,r,\gamma)\) with continuous state and action spaces.  
The policy operates on a state vector at time $t$, \(\mathbf{s}_{t} =[\mathbf{q}^{-}_t, {\dot{\mathbf{q}}}_t^{r}, \phi_t] \) constructed from the simulated environment, consisting of 
the generalized coordinates excluding the root global horizontal position $\mathbf{q}^{-}_t$, 
generalized velocities expressed in the root heading-aligned frame ${\dot{\mathbf{q}}}_t^{r} $, and
the normalized gait phase \(\phi_{t} \in [0,1]\). 
The action \(\mathbf{a}_{t} = [\, \boldsymbol{\psi}^{eq}_{t},\; \boldsymbol{\xi}_{t} ]^{\top}
\) concatenates  equilibrium joint angles $\boldsymbol{\psi}^{eq}_{t}$ and residual wrench $ \boldsymbol{\xi}_{t}$.  
The transition to the next state \(\mathbf{s}_{t+1}\) is governed by the MuJoCo environment, simulating the dynamics (\cref{eq:fd}) at 450 Hz over the policy's control interval \(\Delta t = 1/30\) s, and 
the joint moments \(\boldsymbol{\tau}\) are generated by the impedance control law, 
\begin{equation}
   \label{eq:impedance}
   \boldsymbol{\tau} = -\mathbf{K}_{p}\big( \boldsymbol{\psi} - \boldsymbol{\psi}^{eq}\big) - \mathbf{K}_{d}\,\dot{\boldsymbol{\psi}},
\end{equation}
where $\mathbf{K}_{p}$ and $\mathbf{K}_{d}$ are the stiffness and damping coefficients as diagonal matrices. The equilibrium joint position \(\boldsymbol{\psi}^{eq}\) is held constant throughout the control interval $\Delta t$; likewise, the residual wrench \(\boldsymbol{\xi}\) is held constant and applied to the root during the interval. 

\subsubsection{Learning Objective}  
 Given an expert motion clip of duration $T$ with generalized coordinate trajectory \(\{\mathbf{q}^{\texttt{exp}}_{t}\}_{t=0}^{T}\), the corresponding GRFs \(\{\mathbf{f}^{\mathrm{grf}_{\texttt{exp}}}_{t}\}_{t=0}^{T}\) and  CoP profiles \(\{{\mathbf{x}}^{\mathrm{cop}_{\texttt{exp}}}_{t}\}_{t=0}^{T}\), the goal is to learn a stochastic policy \(\pi_{\theta}(\mathbf{a}_{t}\mid\mathbf{s}_{t})\) that maximizes the expected cumulative reward  
 \(
 J(\theta) = \mathbb{E}_{\pi_{\theta}}\big[\sum_{t} \gamma^{t} r\!\left(\mathbf{s}_{t}, \mathbf{a}_{t}, \mathbf{q}^{\texttt{exp}}_{t}, \mathbf{f}^{\mathrm{grf}_{\texttt{exp}}}_{t},
\mathbf{x}^{\mathrm{cop}_{\texttt{exp}}}_{t}\right)\big],
 \)
where \(\gamma\) is the discount factor and \(\theta\) denotes the network parameters. The reward \(r(\cdot)\) penalizes deviations between learned and expert kinematics, such as joint poses, end-effector positions, and root motion in MOIL. We additionally penalized deviations from the expert kinetic signals in KAIL, as further explained in \cref{subsec:rew_design}.  

\subsection{Reward Design}
\label{subsec:rew_design}
\begin{figure}
    \centering
    \includegraphics[width=\linewidth]{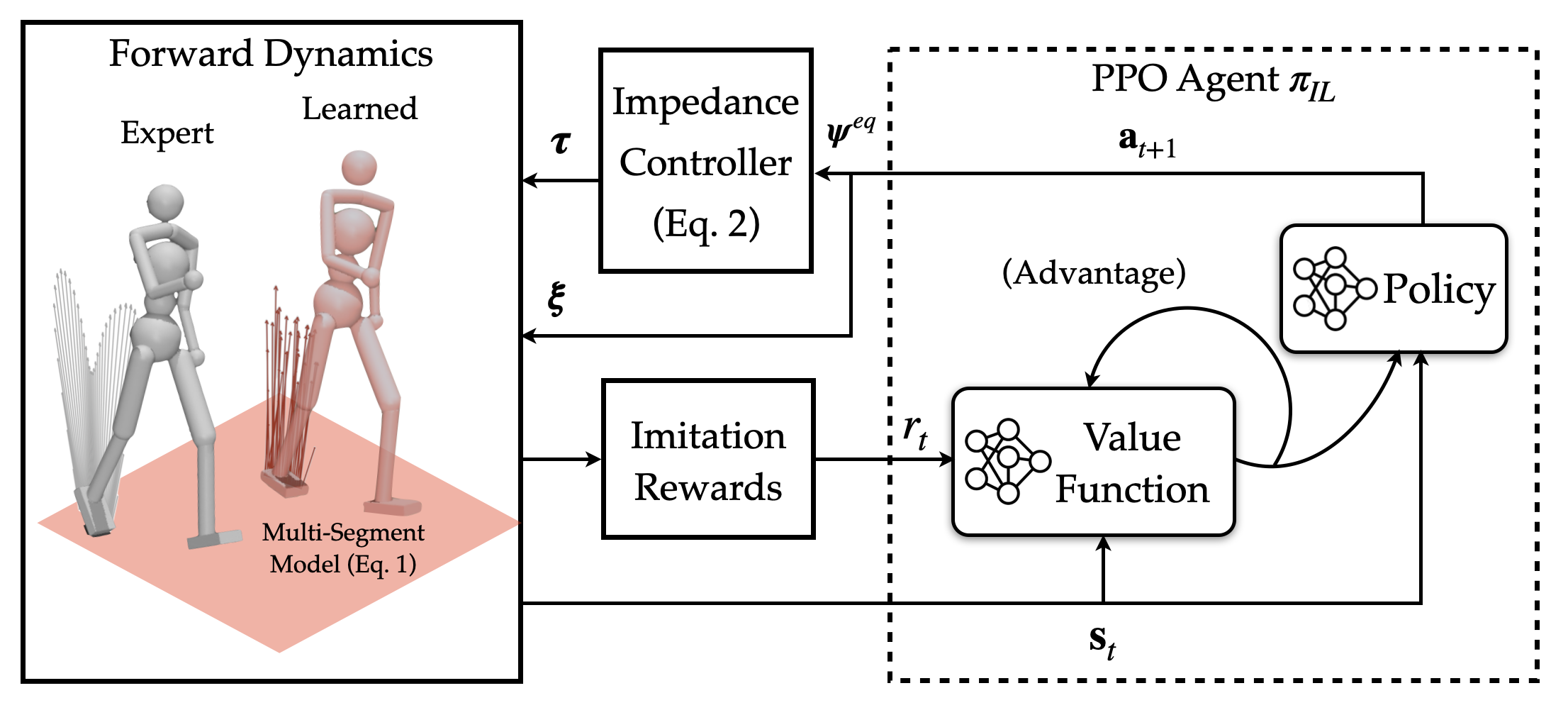}
    \caption{
    {Detailed biomechanics imitation-learning pipeline corresponding to the dashed box in \cref{fig:intro}(b).  The proximal policy optimization (PPO) agent learns to imitate expert gait from rewards defined in \cref{subsec:rew_design}, while the impedance controller and MuJoCo-based multi-segment model generate the forward simulation in response to the agent’s actions.}
    }

\label{fig:ppo_schematic}

    \label{fig:method}
\end{figure}

The reward is decomposed {into kinematics-related terms, aggregated as $R^\mathrm{k}$, and additional kinetics-related terms 
that encourage agreement with the expert’s kinetic measurements. Specifically, $R^\mathrm{k}$ is defined as the weighted sum of multiple kinematic imitation objectives, each expressed using a negative-exponential reward:}
\begin{equation}
\label{eq:rew_kinematics}
      R^\mathrm{k} = w^\mathrm{p} R^\mathrm{p} + w^\mathrm{ee} R^{\mathrm{ee}} + w^{\mathrm{rp}} R^{\mathrm{rp}} + w^{\mathrm{rv}} R^{\mathrm{rv}} + w^{\mathrm{vf}} R^{\mathrm{vf}}.
\end{equation}
 These terms, with their respective weights, encourage agreement with the expert’s joint kinematics ($R^{\mathrm{p}}$), end‑effector positions ($R^{\mathrm{ee}}$), root height and orientation ($R^{\mathrm{rp}}$), as well as root linear and angular velocities ($R^{\mathrm{rv}}$). A residual‑wrench regularization term ($R^{\mathrm{vf}}$) penalizes excessive corrective wrench $\boldsymbol{\xi}_{t}$, promoting dynamically feasible motion. 

We further enforce \textit{kinetics-aware} imitation by introducing kinetics-related reward terms 
as defined below.
\subsubsection{Ground Reaction Force (GRF) reward} 
    \begin{equation}
    \label{eq:rew_grf}
    \begin{aligned}
        R^{\mathrm{grf}} = \exp\!\Big(&
        -\alpha^{\mathrm{grf}} 
        \left\| \Delta \mathbf{f}^{\mathrm{grf}} \right\|^{2}  
         \Big),
    \end{aligned}
    \end{equation}

Here, $\alpha^{\mathrm{grf}} $  controls the decay rate of the negative-exponential term, and $\Delta \mathbf{f}^{\mathrm{grf}}$ 
stacks the difference of sagittal plane GRFs of learned interaction forces with the expert profiles from treadmill measurement, 
\[
\Delta \mathbf{f}^{\mathrm{grf}} =
[f^{ap}_{r} - f^{ap_{\texttt{exp}}}_{r}, \\
f^{ap}_{l} - f^{ap_{\texttt{exp}}}_{l}, \\
f^{v}_{r} - f^{v_{\texttt{exp}}}_{r}, \\
f^{v}_{l} - f^{v_{\texttt{exp}}}_{l}] ^{\top},
\]

where each term is the difference between learned and expert anterior-posterior (\textit{ap}) or vertical (\textit{v}) GRFs for the right (\textit{r}) and left (\textit{l}) feet at each time step. 

\subsubsection {Center of Pressure (CoP) reward}
    \begin{equation}
    \label{eq:rew_cop}
        R^{\mathrm{cop}} = \exp\!\Big(
        -\alpha^{\mathrm{cop}} 
        \left\| \Delta \mathbf{x}^{\mathrm{cop}} \right\|^{2}  
         \Big),
    \end{equation}

    where $\alpha^{\mathrm{cop}}$ is the decay rate of the negative-exponential term, and 
    \(\Delta \mathbf{x}^{\mathrm{cop}}= [\,x^{ap}_{r} - x^{ap_{\texttt{exp}}}_{r}, x^{ap}_{l} - x^{ap_{\texttt{exp}}}_{l}\,]^{\top}\)
    represents the difference between learned and expert anterior-posterior CoP position for the right and left feet. 
    
 {The specific parameter values are detailed in \cref{tab:rew}. 
For kinematics-related terms in \cref{eq:rew_kinematics}, the parameters follow Yuan et al.~\cite{yuan_residual_2020}, {except for a heavier penalty on the residual wrench to constrain its magnitude}.  For kinetics-related terms, the corresponding parameters were selected 
{from a parameter sweep based on training progress and fixed before evaluation on the held-out gait cycles.}
    } 


\begin{table}[t] \centering 
\caption{{ Hyperparameters of the kinematics- and kinetics-related reward terms. The coefficient $\alpha$ controls the exponential decay, and the weight 
$w$ scales the corresponding reward term. {The last two rows show aggregated kinematic and kinetic reward weights. Parentheses show the aggregated reward weight percentage.}
}}
\label{tab:rew} \setlength{\tabcolsep}{5pt} 
\begin{adjustbox}{max width=0.8\linewidth} 
\label{tab:rew} 
\setlength{\tabcolsep}{5pt} 
\begin{tabular}{lccc} 
\toprule
\textbf{Reward Term} & \textbf{Symbol} & \textbf{Coef. $\alpha$} & \textbf{Weight $w$} \\ 
\midrule
Pose                           & $R^{\mathrm{p}}$   & 2.0       & 0.50 \\ 
End-Effector                  & $R^{\mathrm{ee}}$  & 20      & 0.10 \\ 
Root Height/Orientation    & $R^{\mathrm{rp}}$  & 100/500 & 0.15 \\ 
Root Lin./Ang. Velocities  & $R^{\mathrm{rv}}$  & 1.0/0.1 & 0.10 \\ 
Residual {Force/Moment}                 & $R^{\mathrm{vf}}$  & {1.0/5.0}      & {0.20}\\
\midrule
Ground Reaction Force & $R^{\mathrm{grf}}$ & {2.0}       & 0.25 \\ 
Center of Pressure      & $R^{\mathrm{cop}}$ & 1.0       & 0.25 \\
\midrule
{Aggr. Kinematics}        & $R^{\mathrm{k}}$  & --       & {1.05 (68\%)} \\
{ Aggr. Kinetics}     & -- & --       & {0.50 (32\%)} \\
\bottomrule 
 \end{tabular} \end{adjustbox} \end{table}

\subsection{Training Setup}  
 {Proximal Policy Optimization (PPO) \cite{ppo2017} was used to train the policy in the aforementioned environment (\cref{fig:method})}. The policy was modeled as 
a Gaussian stochastic actor with fixed action standard deviation \(\exp(-2.3)\), 
implemented as a two-layer ReLU network with 512 and 256 hidden units. The value 
function used an identical architecture. Both networks were optimized using the ADAM optimizer, 
with learning rates of \(5\times10^{-5}\) for the policy and \(10^{-3}\) for the 
critic.
Training employed PPO with clipping parameter \(\epsilon_c=0.2\) 
and generalized advantage estimation with discount factor \(\lambda=0.95\) and generalized advantage estimation mixing parameter \(\tau = 0.95\). Each training iteration 
collected 50,000 environment steps, followed by 10 epochs of mini-batch gradient updates 
with a batch size of 2048. 
Training ran for up to {1100} iterations on an NVIDIA RTX A2000 GPU and an Intel i9-10900X CPU. 

\section{EXPERIMENTS}
\label{sec:experiments}
\subsection{Data Collection and Processing}
{Two healthy individuals (Participant 1 (AB1): age 23 years, height 170 cm, body mass 60 kg; Participant 2 (AB2): age 29 years, height 176 cm, body mass 70 kg) with no history of musculoskeletal, orthopedic, or neurological disorders participated in this study.} The study protocol was reviewed and approved by the institutional review board 
(IRB No. 24671).

{To generate reference data for training and evaluation, the participant{s} completed three walking trials 0.9, 1.2, 1.5 m/s on a treadmill, each lasting 2 minutes.}
Forty-two reflective markers were placed on anatomical landmarks. Three-dimensional kinematic data were collected from the optical motion capture system (Vicon, Oxford, UK) at a sampling frequency of 120 Hz. GRFs were measured from the force-plate-embedded split-belt treadmill (Bertec, Columbus, OH, USA) at a sampling frequency of 1000 Hz. 


The coordinates of the markers and GRFs data were filtered using a zero-lag fourth-order Butterworth filter with cutoff frequencies of 6 Hz and 10 Hz, respectively. Inverse kinematics and dynamics were performed in Visual3D biomechanical modeling software (HAS-Motion, Ontario, Canada).
{The gait cycle was defined from right heel-strike to the subsequent ipsilateral heel-strike using a 20~N vertical GRF threshold, and phase variable $\phi$ was defined relative to this reference. To ensure steady-state walking, the first 10~s of each trial were excluded. The next 20 strides were used for training, and the subsequent 15 strides for testing.}

\begin{figure}
    \centering
    \includegraphics[width=\linewidth]{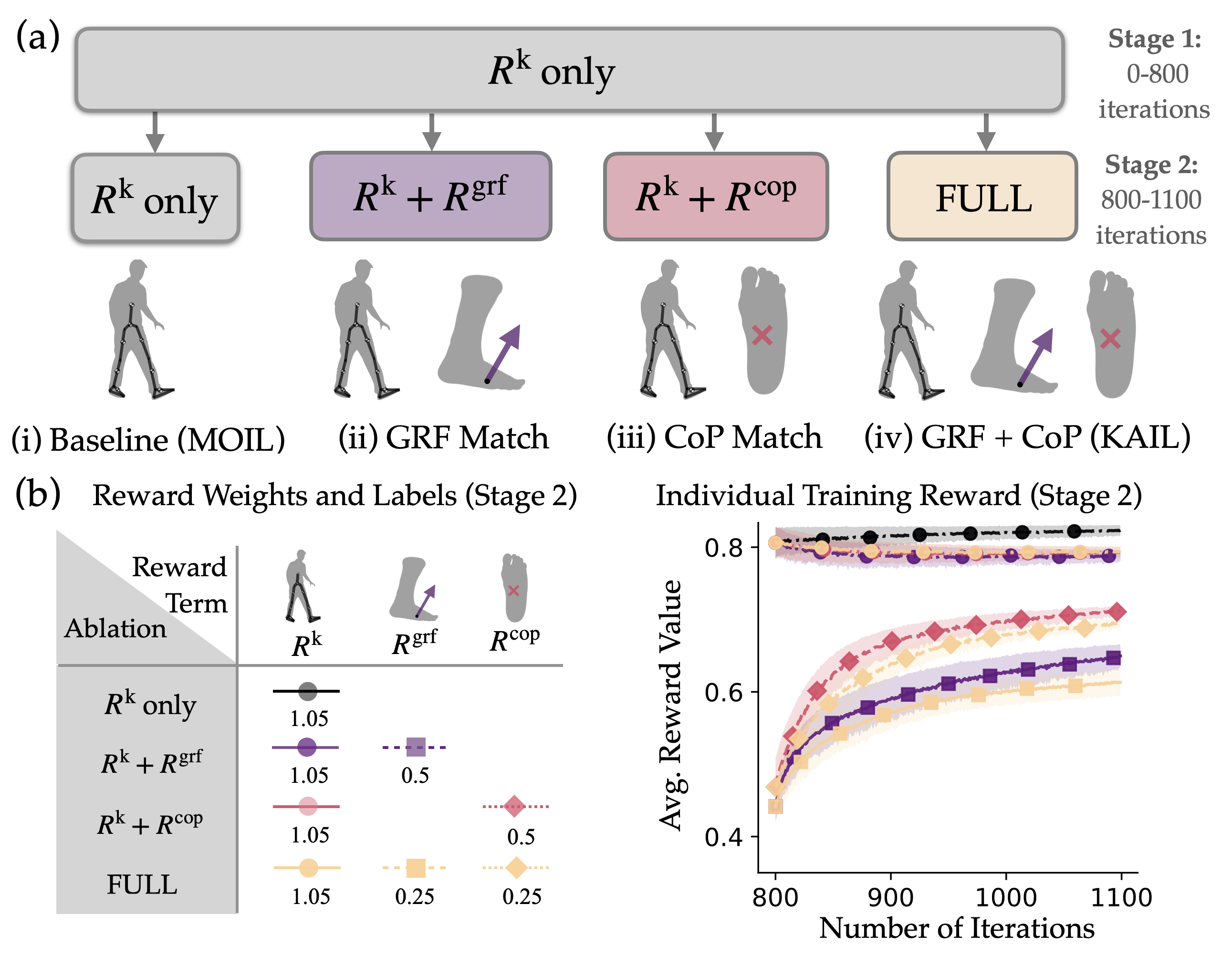}
     \caption{Progressive Kinetic Reward Shaping scheme and corresponding reward curves. {(a) Schematic of the training protocol, where agents are pre-trained for {800} iterations with kinematic reward only ($R^{\mathrm{k}}$) before entering a 300-iteration fine-tuning phase, resulting in four shaping conditions (i)--(iv).}
     (b) {Individual training reward curves of each active component during the second stage{ for AB1, averaged across five random seeds}. The left table lists the line colors, markers, and corresponding reward weights for each progressive kinetic rewarding shaping condition and reward term. Empty entries indicate inactive rewards with zero weight. The weight of $R^{\mathrm{k}}$ ({1.05}) is the sum of the weights of all kinematics-related reward terms in \cref{eq:rew_kinematics} according to \cref{tab:rew},  and $R^{\mathrm{k}}$ reward value is normalized by this equivalent weight for comparability.}
     }
    \label{fig:reward_curriculum}
\end{figure}

\subsection{Imitation Learning via Progressive Kinetic Reward Shaping Design}
{We employed a progressive, two-stage, kinetic reward shaping approach to evaluate the contribution of the additional reward components, as shown in \cref{fig:reward_curriculum} (a). 
Specifically, all experiments shared a common baseline policy trained with kinematics reward terms for the first {800} iterations. Starting from this identical checkpoint, training was continued for an additional 300 iterations under four reward configurations{, each trained with five random seeds}: (i) kinematics only (\textbf{Baseline/MOIL}), (ii) kinematics+GRF (\textbf{GRF Match}), (iii) kinematics+CoP (\textbf{CoP Match}), and (iv) kinematics+GRF+CoP (\textbf{KAIL}). {Reward weights were held constant for each stage and condition sduring training. }
This experimental design isolates the incremental contribution of GRF and CoP while controlling for differences in initialization and early-stage learning. {Unless otherwise stated, metrics were first averaged across seeds, then across walking speeds, and finally across participants.}
}
{
The detailed reward weight schedule is shown in \cref{fig:reward_curriculum}(b). For comparability, {the share of the total reward weight assigned to kinematic tracking terms} was kept constant in conditions (ii)--(iv). 
For {each participants and} speed, we completed the training protocol to obtain the corresponding policies for evaluation.}

\subsection{Model Evaluation and Analysis}
\label{subsec:exp_eval}


{We evaluated the kinematic accuracy, external kinetics fidelity, and internal kinetics plausibility of the four conditions using the following outcome signals: joint angles, 
GRFs, CoP trajectories, and joint moments.}

{For each progressive kinetic reward shaping condition and walking speed, the trained policy was rolled out in forward simulation for evaluation. The corresponding signals were segmented into gait cycles using the experimental heel-strike events. 
GRFs were projected onto the sagittal plane.
CoP trajectories were computed as the vertical GRF-weighted average of contact locations in the foot frame and normalized by foot length. Joint moments were normalized by body mass. All signals were filtered with a fourth-order 6~Hz Butterworth filter and linearly interpolated to 101 points over the gait cycle (0--100\%)}. 
{Root mean square error (RMSE) was then computed for joint angles, CoM velocities, GRFs, CoP trajectories, and joint moments relative to the corresponding experimental references (inverse kinematics, force-plate measurements, and inverse dynamics) within each gait cycle from the test dataset.}

\section{RESULTS}
\label{sec:results}

\begin{figure}

    \centering
    \includegraphics[width=\linewidth]{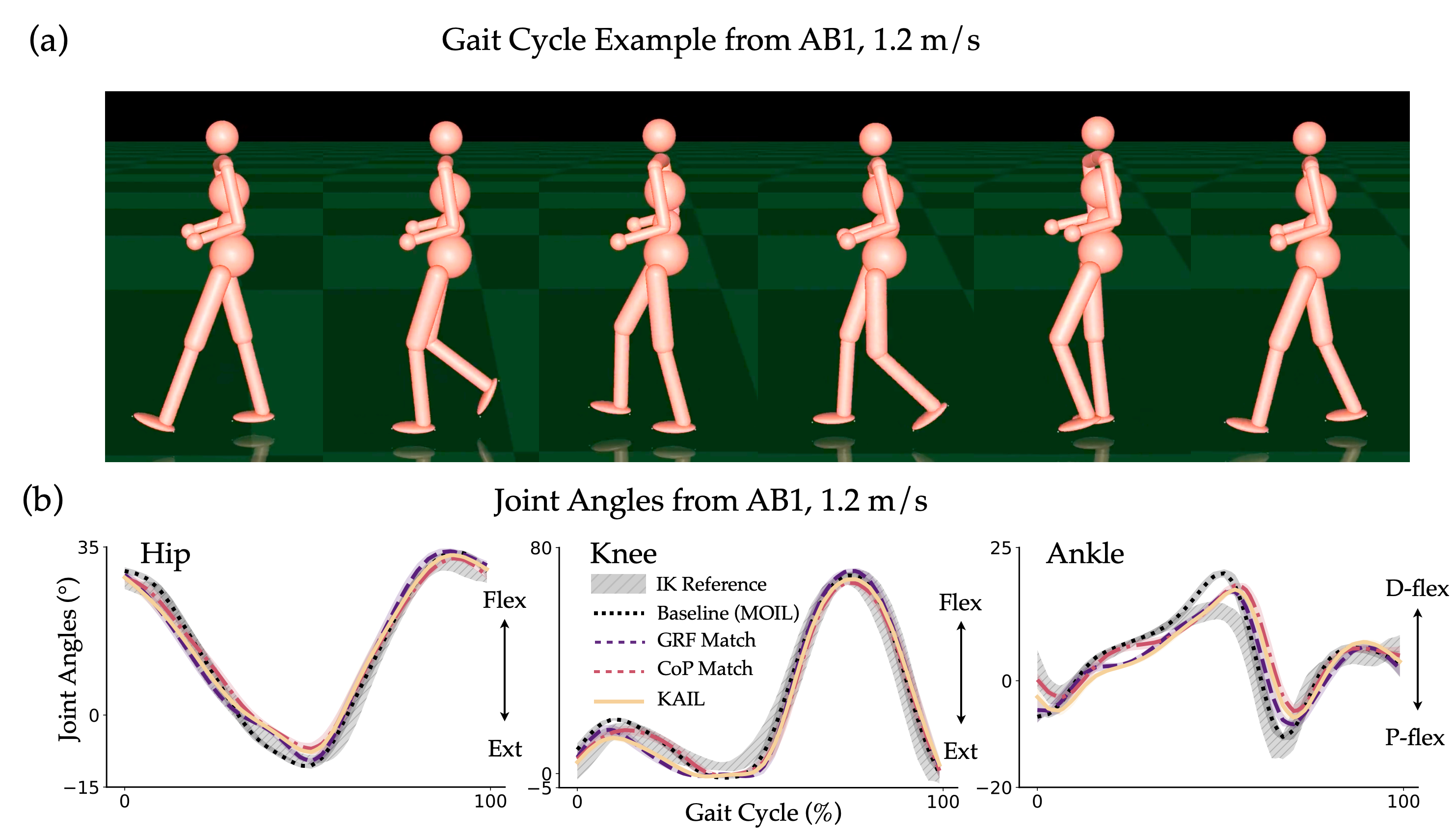}
    \caption{
    {(a) MuJoCo motion clip of one gait cycle from AB1, 1.2 m/s.
    (b) Mean hip, knee, and ankle angles over the gait cycle for the four progressive kinetic reward shaping conditions, compared to inverse kinematics from human experiments (mean $\pm$ 2 standard deviations); Flex/Ext: flexion/extension;  D-flex/P-flex: ankle dorsiflexion/plantarflexion.    }}
     

    \label{fig:angles}
\end{figure}

\begin{figure*}
    \centering
    \includegraphics[width=\linewidth]{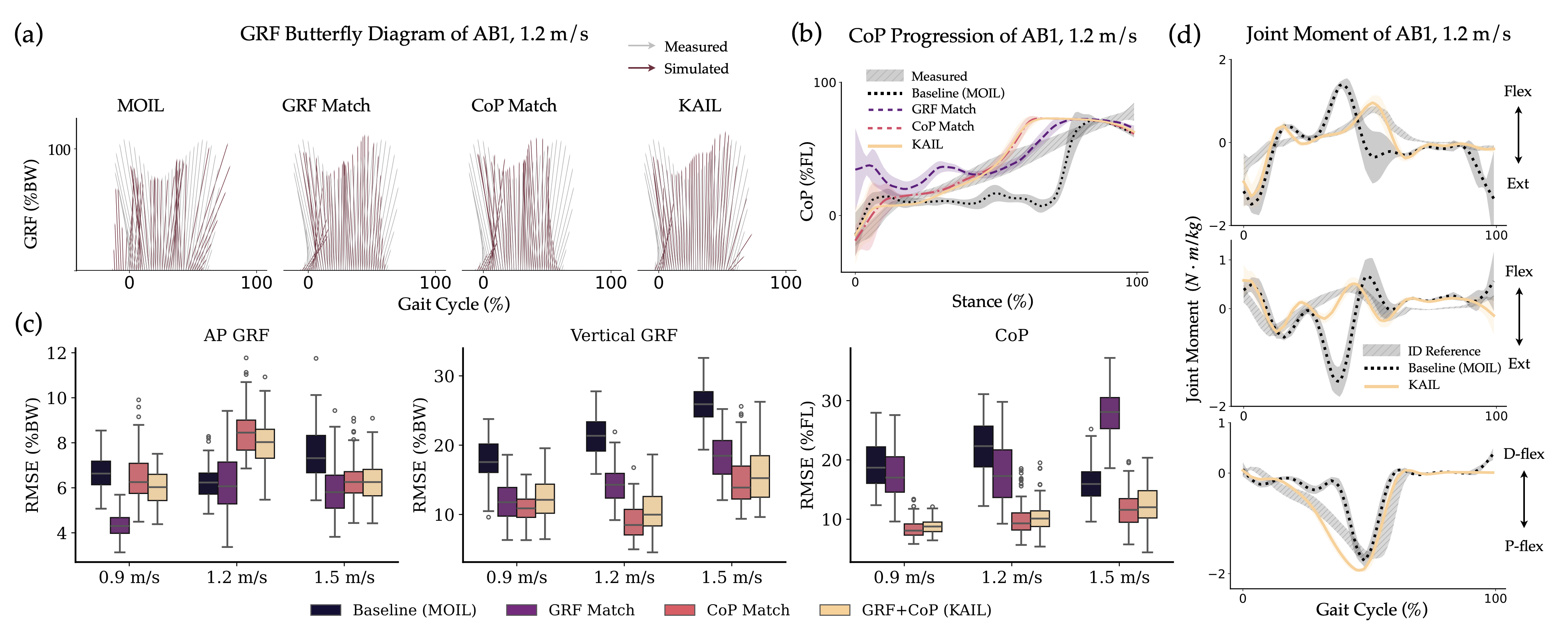}
    \caption{(a) Sagittal GRF butterfly diagrams of AB1 at 1.2 m/s: forward-simulated vs. force-plate measurements across kinetic reward shaping conditions, in \% body weight (\% BW). 
(b) Stance CoP progression of AB1 at 1.2 m/s across kinetic reward shaping conditions, in \% foot length (\% FL), vs. force-plate measurements (mean $\pm$ 2 standard deviations).
(c) IQR box plots of GRF RMSE (anterior–posterior and vertical) and CoP RMSE across speeds and reward shaping conditions for AB1.
(d) Mean forward-simulated joint moment trajectories over the gait cycle vs. inverse dynamics reference (plotted with mean $\pm$ 2 standard deviations) across conditions for AB1 at 1.2 m/s.}
    \label{fig:grfs}
\end{figure*}

{Results are reported at two levels: the full four-condition ablation (MOIL, GRF Match, CoP Match, and KAIL) is presented for the representative participant AB1 (Fig.~\ref{fig:reward_curriculum}--\ref{fig:grfs}), whereas MOIL and KAIL are compared across $n=2$ participants (\cref{tab:kinematics} and \cref{fig:stats}).}

\subsection{Learning Performance}
{ During the first stage, MOIL pretraining produced stable walking policies for all walking speeds. The subsequent fine-tuning stage improved the targeted kinetic rewards while preserving high kinematic performance.  The progressive kinetic reward shaping approach was conducted to evaluate the contribution of each kinetics-aware reward term to policy refinement. Across walking speeds{, participants and random seeds}, all conditions completed the full walking trials without falls. {For AB1 (\cref{fig:reward_curriculum}(b)),} the aggregated kinematic reward $R^\mathrm{k}$ remained consistently within {0.78--0.82} (circles) regardless of the addition of kinetics-aware rewards.}
{In the GRF Match condition, the GRF reward increased from {0.45 to 0.65},
and the CoP reward increased from {0.48 to 0.71} in the CoP Match condition.
In the KAIL condition, the GRF and CoP rewards increased from {0.45 to 0.61} and from {0.48 to 0.69}, respectively.
}

\subsection{Kinematics Fidelity}
\label{subsec:res_kinematics}
\begin{table}[t]
\centering
\caption{{Kinematic tracking error and mean residual wrench for MOIL and KAIL. Values are averaged across five random seeds, three walking speeds and two participants. $F^{v}_{\mathrm{res}}$ and $F^{ap}_{\mathrm{res}}$ denote the vertical and anterior-posterior residual forces, and $M^{sag}_{\mathrm{res}}$ the sagittal-plane residual moment. 
}}
\label{tab:kinematics}
\footnotesize
\setlength{\tabcolsep}{4.5pt}
\begin{tabular}{lcccccc}
\toprule
& \multicolumn{3}{c}{Kinematics RMSE} & \multicolumn{3}{c}{Residual Wrench RMS} \\
\cmidrule(lr){2-4} \cmidrule(lr){5-7}
Condition & Hip & Knee & Ankle & $F^{v}_{\mathrm{res}}$ & $F^{ap}_{\mathrm{res}}$ & $M^{sag}_{\mathrm{res}}$ \\
& ($^\circ$) & ($^\circ$) & ($^\circ$) & (\% BW) & (\% BW) & (N$\cdot$m/kg) \\
\midrule
MOIL & 2.26 & 4.18 & 3.24  & 0.73\rlap{$^{*}$} & 0.51\rlap{$^{*}$} & 0.009\rlap{$^{*}$} \\
KAIL & 3.58 & 5.35 & 4.82  & 2.86\rlap{$^{*}$} & 0.84\rlap{$^{*}$} & 0.042\rlap{$^{*}$} \\
\bottomrule
\end{tabular}

\vspace{2pt}
\begin{minipage}{\linewidth}\footnotesize\raggedright
$^{*}$Below residual threshold~\cite{hicks2015my}
$F_{res} < 5\% ~\max[\mathbf{f}^\mathrm{grf}] \approx 6\% BW$ for forces,  $M_{res} < h^{CoM} ~\max[\mathbf{f}^\mathrm{grf}/BW] \approx 0.12$~N$\cdot$m/kg  for moments.
\end{minipage}
\end{table}

\cref{fig:angles}{(b)} shows the average hip, knee, and ankle joint angle profiles for the learned kinematics {at 1.2 m/s}, along with inverse kinematics joint angle profiles.  {\cref{tab:kinematics} summarizes the joint angle tracking errors across participants and walking speeds. MOIL and KAIL achieved comparable kinematic tracking, 
with mean joint angle RMSEs differing by less than 1.6° at each joint. 
}
%
{Across participants, the sagittal plane residual forces and moments remained below the dynamic-consistency thresholds recommended by Hicks et al.~\cite{hicks2015my} in both conditions, although they were larger for KAIL (\cref{tab:kinematics}).}

\begin{figure}
    \centering
    \includegraphics[width=\linewidth]{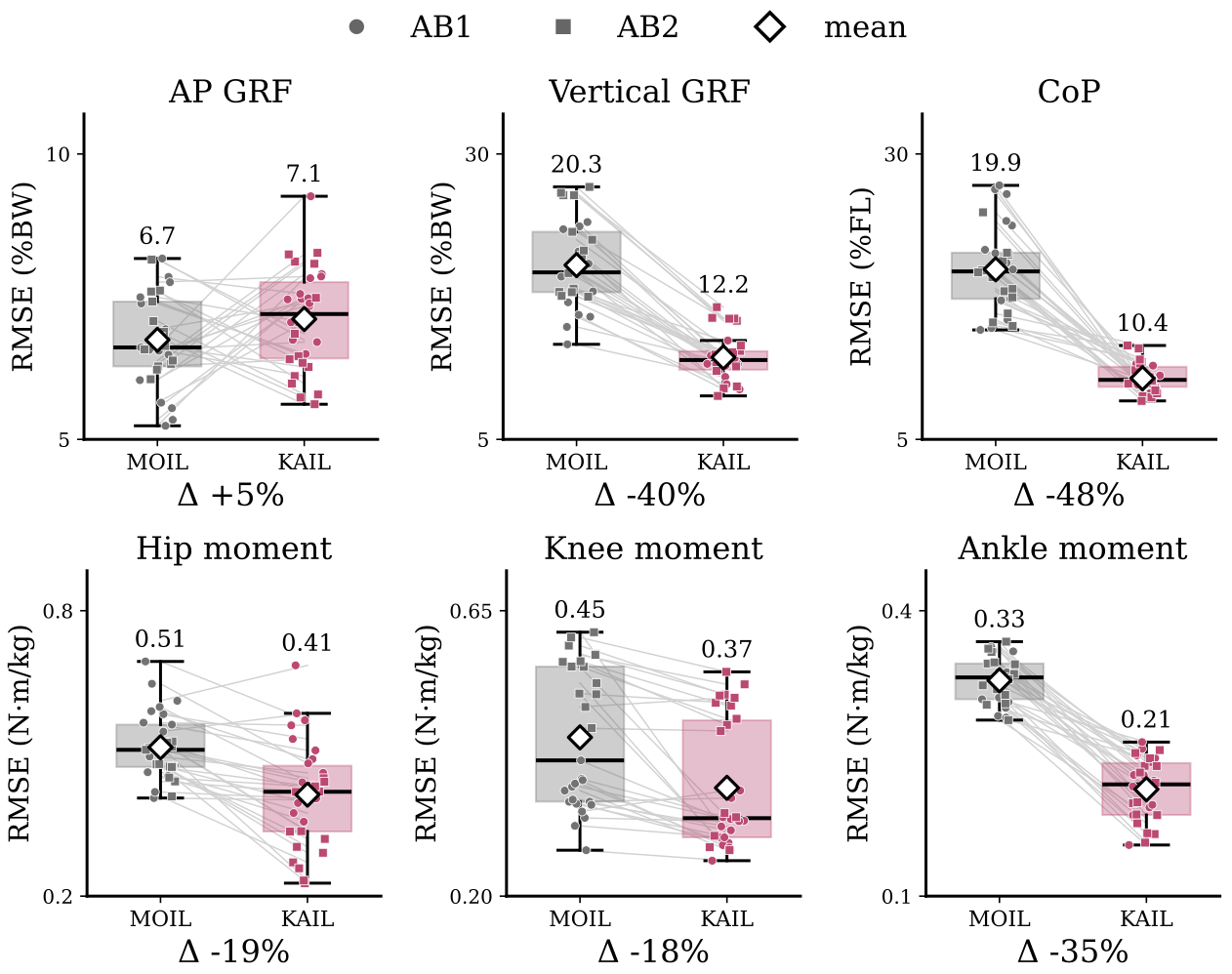}
    \caption{{RMSE of the AP GRF, vertical GRF, CoP, and hip, knee, and ankle joint moments for MOIL and KAIL, relative to the experimental measurements and inverse dynamics reference. Each point is one training run (two participants, three walking speeds, and five random seeds; $n = 30$ per condition), and gray lines connect the paired MOIL and KAIL runs. $\Delta$ denotes the relative change in mean RMSE from MOIL to KAIL.}}
    \label{fig:stats}
\end{figure}

\subsection{External Kinetics Fidelity}

\label{subsec:res_kinetics}
\label{subsec:res_kinetics}
{The average GRF profiles and CoP trajectories along with experimental measurements {of AB1 }at 1.2 m/s are shown in \cref {fig:grfs}(a) and (b), respectively. For each shaping condition and walking speed, external kinetics error was quantified using the RMSE between simulated and measured GRFs and CoP locations, as shown in \cref{fig:grfs}(c).}

{
{The kinetics-related rewards affected the two GRF directions differently (\cref{fig:grfs}(c)): the vertical GRF RMSE decreased most clearly in the CoP Match and KAIL conditions, whereas the AP GRF RMSE decreased only in the GRF Match condition and was comparable to or higher than that of MOIL in the CoP Match and KAIL conditions.} Across three speeds {for AB1 (\cref{fig:grfs}~(c)),
GRF Match, CoP Match, and KAIL reduced vertical GRF RMSE by 32 ± 2\%, 46 ± 10\%, and 40 ± 10\%, respectively, relative to MOIL. In contrast, GRF Match reduced AP GRF RMSE by 18 ± 18\%, whereas CoP Match and KAIL did not improve it. Across both participants and all three speeds, KAIL reduced the mean vertical GRF RMSE by 40\%, while increasing the mean AP GRF RMSE by 5\% relative to MOIL (\cref{fig:stats})}.

{CoP regulation produced more consistent reductions in CoP RMSE
than GRF matching alone (\cref{fig:grfs}(c)).
For AB1 across speeds, CoP Match and KAIL reduced CoP
RMSE by 44 $\pm$ 17\% and 43 $\pm$ 16\%, respectively,
relative to MOIL.
Both conditions reduced CoP RMSE at all three speeds, whereas
GRF Match reduced CoP RMSE at the two lower speeds but increased
it at the highest speed.
Across both participants and all three speeds, KAIL reduced the
mean CoP RMSE by 48\% relative to MOIL (\cref{fig:stats}).}

\subsection{Internal Kinetics Plausibility}

{{\cref{fig:grfs}(d)} shows average simulated joint moment profiles against the inverse dynamics reference {of AB1 at 1.2~m/s}. Importantly, none of the conditions were trained with inverse dynamics moments as explicit supervision; these references were used only for post hoc evaluation. 
Qualitatively, MOIL yielded irregular estimates at key gait events, particularly at heel-strike and {late stance}. In contrast, KAIL produced smoother profiles that more closely aligned with reference inverse dynamics outcomes, notably at the {ankle}.}

{Joint moment accuracy was evaluated
for the MOIL and KAIL conditions at each walking speed 
using RMSE relative to the inverse dynamics reference moments ({\cref{fig:stats}}). 
{Across two participants, three speeds and five seeds, KAIL reduced the joint moment RMSE at the hip (by 19\%), knee (by 18\%), and ankle (by 35\%).}
}

\section{DISCUSSION}
\label{sec:discussion}

This study examined whether MOIL is sufficient to produce biomechanically plausible joint moments for forward simulation of human gait. The results show that although MOIL reproduced human gait kinematics, it did not reliably produce human-like joint moments or foot-ground interactions. Since joint moment calculation in walking is sensitive to foot contact force and location during the stance phase \cite{bj2026}, when we added the measurements of GRF and CoP location into the reward function during training (i.e., KAIL), the policy yielded {joint moments with RMSE reduced by 19\% at the hip, 18\% at the knee, and 35\% at the ankle across participants, in forward simulations of walking, compared to those derived from MOIL.}
This implies that MOIL does not guarantee the production of human-like kinetics in gait simulations, whereas KAIL yields more biomechanically plausible internal and external kinetics.
{Note that} the kinematic errors of both MOIL and KAIL fall within the range reported in the literature \cite{cotton_kintwin_2025, chiu2025speedadaptive}; 
{the residual wrench magnitudes also remained below the dynamic-consistency threshold \cite{hicks2015my} and were substantially smaller than the differences in moment estimates. Therefore, neither kinematic errors nor residual wrenches can explain the performance gap between MOIL and KAIL. 
}



{
MOIL consistently exhibited non-physiological GRF and CoP patterns (\cref{fig:grfs}).
Specifically, the discrepancies were characterized by premature foot contact, prolonged stance, attenuated heel-strike GRF peaks, and a {heel}-biased CoP trajectory (\cref{fig:grfs}(a-b)). 
These errors propagated to joint moments {at all three joints (\cref{fig:stats})}. 
 The GRF Match condition improved force magnitude and timing, but it still biased the contact locations {(\cref{fig:grfs}(b))}, indicating that GRF constraints alone are insufficient to recover dynamically consistent locomotion. 
 The CoP Match condition complementarily constrained the progression of the force application point under the foot, 
 thereby regularizing the external moment arm and secondarily improving GRF regulation.
Combined, the KAIL formulation generated a more balanced and biomechanically plausible solution. {It substantially reduced the vertical GRF and CoP errors, although not the AP GRF error, and showed closer agreement with inverse dynamics joint moment estimates at all three joints, as indicated by reduced RMSE (\cref{fig:stats}).} 
These results indicate that kinematics alone \textbf{\textit{underconstrain}} the underlying dynamics. Consistent with the identifiability problem in inverse dynamics~\cite{winter2009biomechanics}, 
similar joint motion can arise from different external contact wrenches and joint moments (\cref{eq:fd}); therefore, accurate contact measurements are needed to constrain the the solution space of joint moments further to avoid this non-uniqueness. 
}

{
Despite improved kinetic estimation with KAIL, the generated joint moments still differed from the inverse dynamics results, indicating model mismatch. 
  Trade-offs between kinetic objectives were also observed (\cref{fig:grfs}{(c)}, \cref{subsec:res_kinetics}), 
 as GRF and CoP impose distinct and sometimes competing physical reward objectives; thus, joint optimization does not guarantee improvement in every metric.
These discrepancies may first be attributed to the simplified foot-ground interaction model. 
The simplified geometry in our MuJoCo foot model without a metatarsophalangeal joint, and suboptimal foot-ground contact configurations, likely constrained accurate reproduction of GRFs and CoP profiles~\cite{saraiva2022review}, as evidenced in \cref{fig:grfs}(a) and (b). These errors can affect more proximal joint kinetics, as prior work has shown that knee and hip moments are particularly sensitive to uncertainty in CoP location~\cite{camargo2013influence}. 
Another source of error may arise from the non-biological actuation dynamics of the dynamic model (\cref{subsec:dynamics}, \cref{eq:impedance}). The model used an impedance controller with preset stiffness and damping coefficients, which may yield jerkier joint moment profiles with higher-frequency content. In contrast, biological joint moments are generated through the musculoskeletal structure, where muscle behaves as a low-pass actuator~\cite{vivo1979quantitative}. Therefore, even when external kinetics are encouraged through reward terms, the resulting joint moments may retain mechanical features that differ from biological kinetics. Combined with the nonconvex nature of reinforcement learning optimization, these factors can lead to residual kinetics errors and trade-off effects.
}

{By no means do we undermine the potential of IL for human biomechanics. IL-learned control policies remain valuable for their ability to generate stable and continuous human movements in forward dynamics simulations. Such stable walking simulations can be used for cause-and-effect analysis and initial wearable robot control design. Nevertheless, our findings call for a more cautious interpretation of the recent enthusiasm surrounding MOIL for building “digital twin” models of human biomechanics, supporting clinical diagnosis, and enabling the sim-to-real design of wearable robots. Our results indicate that the underlying assumption that MOIL can produce a physiologically plausible model of human gait may not be completely valid. We therefore advocate that IL still requires critical information about foot-ground interactions to explain valid human gait biomechanics. 
}

A few limitations should be acknowledged. First, only lower limb kinematics were considered in this study. The model was simplified to focus on lower limb joint biomechanics, consistent with common inverse dynamics approaches in gait biomechanics. Therefore, excluding upper body kinematics was not expected to compromise the main findings.
{Second, this study was designed to demonstrate through counterexamples how IL can be misapplied in gait kinetics estimation; it was not intended to establish the general benefit of KAIL. Our findings on KAIL are therefore limited to healthy young adults during level walking. Future work should examine whether these findings extend to other populations and whether improving model validity, for example through more realistic foot–ground contact modeling, improves estimation accuracy.
}

\printbibliography

\end{document}